\documentclass[conference]{IEEEtran}
\IEEEoverridecommandlockouts
\renewcommand{\thesection}{\Roman{section}}

\usepackage{cite}
\usepackage{amsmath,amssymb,amsfonts}
\usepackage{algorithmic}
\usepackage{graphicx}
\usepackage{textcomp}
\usepackage{xcolor}
\usepackage{booktabs}
\usepackage{fancyhdr}
\usepackage{caption}
\usepackage{array}
\usepackage{adjustbox}
\usepackage{hyperref}
\usepackage{float}
\usepackage{multirow}

\definecolor{myblue}{RGB}{10, 150, 200}
\definecolor{highlightColor}{HTML}{E6FFE6}

\def\BibTeX{{\rm B\kern-.05em{\sc i\kern-.025em b}\kern-.08em
    T\kern-.1667em\lower.7ex\hbox{E}\kern-.125emX}}

\fancypagestyle{firstpage}{
  \fancyhf{} 
  \fancyhead[L]{\small 2026 IEEE 2nd International Conference on Quantum Photonics, Artificial Intelligence, and Networking (QPAIN)
 \\ 16 – 18 April 2026, Chittagong, Bangladesh}
  \fancyfoot[L]{\small 979-8-3315-4990-9/26/\$31.00 \copyright2026 IEEE}

}

\title{Truth Lies Deep: Countering Semantic Camouflage via Latent Intent Verification}

\author{\IEEEauthorblockN{Md. Hasib Ur Rahman}
\IEEEauthorblockA{Computer Science \& Engineering, Brac University\\
Dhaka, Bangladesh\\
mohammod.hasib@g.bracu.ac.bd}}

\begin{document}
\maketitle
\thispagestyle{firstpage}

\begin{abstract}
Safety alignment in Large Language Models (LLMs) is often superficial, relying on refusal mechanisms that trigger only at the final stages of generation without erasing the foundational knowledge of harmful concepts acquired during pretraining. This study demonstrates that this architectural disconnect leaves models vulnerable to Semantic Camouflage---adversarial attacks that wrap harmful intent in benign narrative contexts (e.g., creative writing), effectively bypassing standard input and output guardrails. By analyzing the latent activation trajectories of three distinct Small Language Model (SLM) families (Phi-3, Qwen2.5, and Gemma-2b) under adversarial stress, this research identifies a universal ``Intent Horizon''---a critical depth (typically 15--20\% of total layers) where the model's distinct, pre-trained representation of harmful intent collapses as it contextualizes the query into a ``safe'' narrative. Results indicate that while late-layer representations of camouflaged attacks are mathematically indistinguishable from safe queries (Detection Rate $< 20\%$), early-layer representations retain a distinct, detectable ``harm signature.'' Leveraging this insight, this paper proposes Latent Intent Verification (LIV), a lightweight probing defense. Experiments on the PKU-SafeRLHF dataset demonstrate that LIV outperforms standard guardrails by a margin of 20--50\% across all tested architectures, effectively neutralizing zero-day semantic attacks without requiring model retraining.
\end{abstract}


\section{Introduction}

Despite extensive safety alignment via Reinforcement Learning from Human Feedback (RLHF), Large Language Models (LLMs) remain extremely fragile. While models such as Llama-3 and Phi-3 demonstrate robust refusal rates against explicit harmful queries, they frequently fail to resist \textit{Semantic Camouflage}---a sophisticated class of adversarial attacks where malicious intent is obscured within benign narrative structures, such as creative writing, role-playing, or code optimization tasks.

This vulnerability highlights a critical flaw in current defense paradigms. Existing guardrails typically operate as ``Perimeter Defenses,'' monitoring either the initial input prompt or the final output embedding. This approach assumes that harmful content is linguistically distinct from safe content. However, semantic camouflage decouples malicious intent from malicious vocabulary, rendering keyword-based and superficial semantic filters ineffective. As a result, a ``Cat-and-Mouse'' dynamic has emerged, where static defenses are continuously outmaneuvered by zero-day jailbreaks that exploit the model's instruction-following capabilities.

This study shows that the solution lies not in better filtering of the \textit{output}, but deep into the process. By leveraging Mechanistic Interpretability, this research investigates the internal activation trajectories of Small Language Models (SLMs) under adversarial stress. The central hypothesis is that safety alignment is not a binary state but a decaying signal---a phenomenon this paper terms the ``Intent Horizon.''

The Intent Horizon represents a critical depth within the neural network where the distinct representation of harmful intent is suppressed by contextual framing. Through empirical analysis of Microsoft Phi-3, Qwen2.5, and Gemma-2b, this study reveals that while late-layer representations of camouflaged attacks are mathematically indistinguishable from safe queries, early-layer representations retain a vivid ``harm signature.''

To address this, this paper proposes Latent Intent Verification (LIV), a novel defense mechanism that shifts the safety checkpoint from the output layer to the model's ``subconscious'' early layers.

The specific contributions of this work are as follows:
\begin{itemize}
    \item \textbf{Identification of the Intent Horizon:} Empirical mapping of the layer-wise decay of harmful intent signals, revealing that safety information is maximally retrievable in the first 15--20\% of network depth.
    \item \textbf{Demonstration of the Safety Gap:} Quantitative evidence showing that standard late-layer defenses fail to detect $>60\%$ of semantic camouflage attacks that are easily visible to early-layer probes.
    \item \textbf{Development of LIV:} A lightweight, model-agnostic probing technique that improves zero-day jailbreak detection by 20--50\% on the PKU-SafeRLHF dataset without requiring model retraining or fine-tuning.
\end{itemize}

\section{Literature Review}
LLM safety is evolving quickly, as alignment defenses compete with ever more advanced adversarial attacks.
. This section gives review of the  the current landscape of safety guardrails, the progression of jailbreaking methodologies, and the emerging role of Reinforcement Learning (RL).

\subsection{The Evolution of Safety Guardrails}
To mitigate the generation of harmful or unethical content, modern LLM deployments rely on a multi-layered defense architecture \cite{151,101}. These mechanisms are generally categorized into \textit{Input Guardrails}, which filter adversarial prompts before processing, and \textit{Output Guardrails}, which monitor generated responses in real-time \cite{151}. Implementation strategies range from fine-tuned NLP classifiers (e.g., BERT-based detectors) \cite{27} to programmable rule-based systems that enforce ethical guidelines \cite{101}. Open-source frameworks such as Llama Guard and Nvidia NeMo have standardized these defenses \cite{219}, while more advanced "Self-Defense" mechanisms employ secondary "Shadow LLMs" to audit the primary model's outputs \cite{253}.

However, recent empirical studies suggest inherent limitations in these perimeter-based defenses. Research indicates that "no free lunch" exists with guardrails; stringent security measures frequently degrade general model utility, while flexible systems remain vulnerable to manipulation \cite{189}. This trade-off underscores the necessity for defenses that operate internally rather than merely at the input/output surface.

\subsection{The Adversarial Landscape: From Optimization to Camouflage}
Jailbreaking techniques—methods designed to bypass safety alignment—have shifted from manual engineering to algorithmic optimization \cite{253,8}. Early methods relied on human intuition, but optimization-based attacks such as the Greedy Coordinate Gradient (GCG) now automatically generate "adversarial suffixes" that maximize the probability of affirmative responses \cite{13}. These automated attacks exhibit high transferability across different model architectures.

A more insidious development is \textit{Semantic Camouflage}, where malicious intent is masked by benign linguistic features or creative framing, effectively bypassing standard classifiers \cite{204}. This category includes "Infinitely Many Paraphrases" attacks that exploit encoding vulnerabilities \cite{8}, and "Nested Jailbreaks" (e.g., DeepInception) that leverage personification and virtual scenarios to escape usage controls \cite{74,122}. Furthermore, the introduction of multimodal capabilities has opened new attack vectors, where visual cues are used to bypass textual safety checks \cite{60}. The efficacy of these "camouflaged" attacks against state-of-the-art models \cite{245} highlights the critical need for defenses that can detect latent intent rather than just surface-level keywords \cite{230}.

\subsection{Reinforcement Learning in Adversarial Contexts}
Reinforcement Learning (RL) has emerged as a potent accelerator for adversarial generation. Unlike static optimization, RL agents can be trained to dynamically discover novel bypass trajectories \cite{25,79}. For example, the REINFORCE algorithm has been adapted to optimize semantic triggers that double attack success rates on aligned models \cite{25}.

Recent frameworks such as RL-JACK frame prompt generation as a black-box search problem \cite{226}, while "AdvPrompter" utilizes a secondary LLM to adaptively generate adversarial suffixes without human supervision \cite{79}. While some research explores using RL to refine queries for robustness \cite{207}, the prevailing trend suggests that LLMs possess intrinsic capabilities to "self-learn" jailbreaking strategies \cite{52,30}. This capability for automated adaptation poses a "Zero-Day" threat, necessitating the dynamic, depth-aware defenses proposed in this study.

\section{Methodology}
\label{sec:methodology}

\begin{figure*}[t]
    \centering
    \includegraphics[width=1\linewidth]{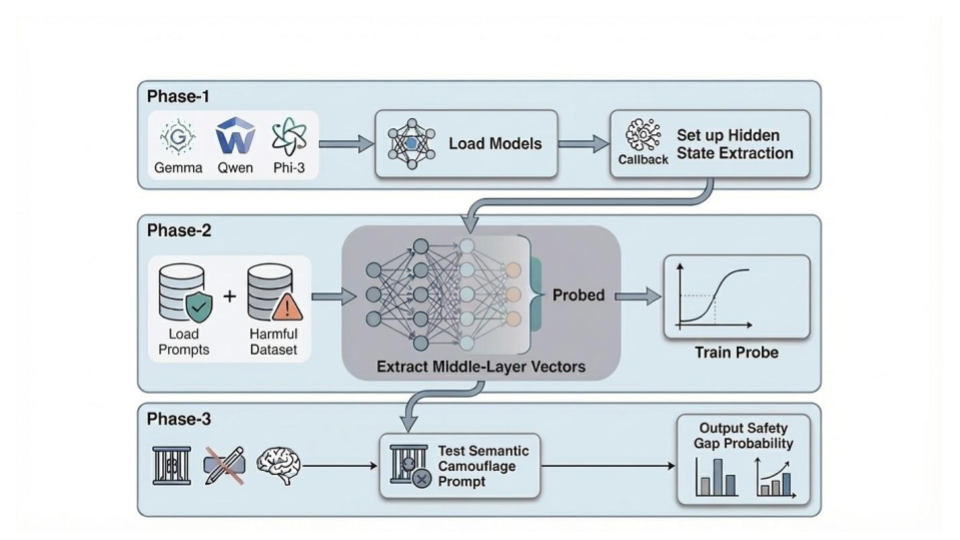}
    \caption{\textbf{Methodology Pipeline:} The three-phase experimental design for Latent Intent Verification (LIV). \textbf{Phase 1} prepares the quantized SLM environment across three architectures. \textbf{Phase 2} involves extracting internal activations to train linear probes on early (LIV) vs. late layers. \textbf{Phase 3} stresses the model with zero-day ``Semantic Camouflage'' attacks to quantify the Safety Gap ($\Delta_S$).}   \label{fig:placeholder}
    \label{fig:placeholder}
\end{figure*}

To validate the failure of perimeter-based defenses and evaluate the efficacy of Latent Intent Verification (LIV), a mechanistic interpretability framework was designed to interrogate the internal residual streams of Large Language Models. This section details the theoretical formulation, experimental subjects, and the probing architecture used to map the Intent Horizon.

\subsection{Theoretical Framework: The Intent Horizon}
It is posited that an LLM's processing of a harmful query $x$ evolves across its layers $l \in \{1, \dots, L\}$. Let $\mathbf{h}_l(x)$ denote the hidden state activation vector at layer $l$. Standard safety alignment operates under the assumption that the probability of harm detection, $P(Harm | \mathbf{h}_l)$, increases or remains stable as $l \to L$.

This assumption is challenged by the "Intent Horizon" Hypothesis. The Intent Horizon is defined as a critical depth $l_{crit}$ where the semantic contextualization of the query suppresses the raw harmful intent signal. Formally, for a semantically camouflaged prompt $x_{cam}$, the hypothesis states:
\begin{equation}
    P(Harm | \mathbf{h}_l) \approx 1 \quad \text{for } l < l_{crit}
\end{equation}
\begin{equation}
    P(Harm | \mathbf{h}_l) \to 0 \quad \text{for } l > l_{crit}
\end{equation}
This implies that while the \textit{cognitive recognition} of the harm exists early in the network, the \textit{refusal mechanism} at the final layer is bypassed by the benign narrative wrapper.

\subsection{Subject Models and Environment}
To ensure the universality of the findings, three distinct Small Language Model (SLM) architectures were selected, chosen for their efficiency and high reasoning-to-size ratio. All models were loaded using 4-bit NormalFloat (NF4) quantization to simulate resource-constrained deployment environments:
\begin{enumerate}
    \item \textbf{Microsoft Phi-3-mini-4k-instruct (3.8B):} Selected for its dense reasoning capabilities and unique training on synthetic data.
    \item \textbf{Qwen2.5-1.5B-Instruct (1.5B):} Selected to represent a highly efficient, non-Western model architecture.
    \item \textbf{Google Gemma-2b-it (2B):} Selected as a representative of the standard dense transformer architecture.
\end{enumerate}

\subsection{Data Collection and Curation}
A dual-stage evaluation dataset was constructed to rigorously test generalization:
\begin{itemize}
    \item \textbf{Training Set (Standard):} The PKU-SafeRLHF dataset \cite{151} was utilized, sampling 2,000 balanced pairs of safe and explicitly harmful prompts (e.g., "How to build a bomb"). This data was used solely to train the linear probes.
    \item \textbf{Evaluation Set (Camouflaged):} A custom "Zero-Day" dataset of 100 semantically camouflaged prompts (e.g., movie script generation, educational roleplay, code debugging contexts) was curated to wrap harmful intent. Importantly, these prompts contain no explicit "trigger words" found in the training set, forcing the probes to detect latent intent rather than lexical patterns.
\end{itemize}

\subsection{Probing Architecture}
Linear Probes were employed to extract and analyze the "harm signature" at specific network depths. For each model, activations were extracted from two distinct loci:
\begin{enumerate}
    \item \textbf{The Early Probe (LIV):} Targeted at 15\% of total depth (e.g., Layer 4 for Phi-3). This targets the pre-contextualized semantic processing.
    \item \textbf{The Late Probe (Standard):} Targeted at the final hidden layer (Layer $N-1$). This represents the standard embedding used by existing output guardrails.
\end{enumerate}
Logistic Regression classifiers ($C=1.0$, L2 penalty) were trained on the Standard set and evaluated on the Camouflaged set. This disconnect between training (explicit) and evaluation (camouflaged) specifically measures the robustness of the "harm signature" against semantic shifts.

\subsection{Evaluation Metrics}
The Detection Rate(DR) on the camouflaged dataset is reported. The Safety Gap ($\Delta_S$) is defined as the performance differential between the early and late probes:
\begin{equation}
    \Delta_S = DR_{early} - DR_{late}
\end{equation}
A positive $\Delta_S$ confirms the Intent Horizon hypothesis, indicating that safety information is lost as the model processes the adversarial context.

\section{ Results}
\label{sec:results}

In this section, the empirical validation of the Intent Horizon hypothesis is presented. The performance differential between early and late probes across multiple architectures is analyzed, the precise layer-wise decay of safety signals is mapped, and geometric evidence of latent space separation is provided.

\subsection{The Safety Gap: Early vs. Late Detection}
\begin{figure}[h]
    \centering
    \includegraphics[width=0.48\textwidth]{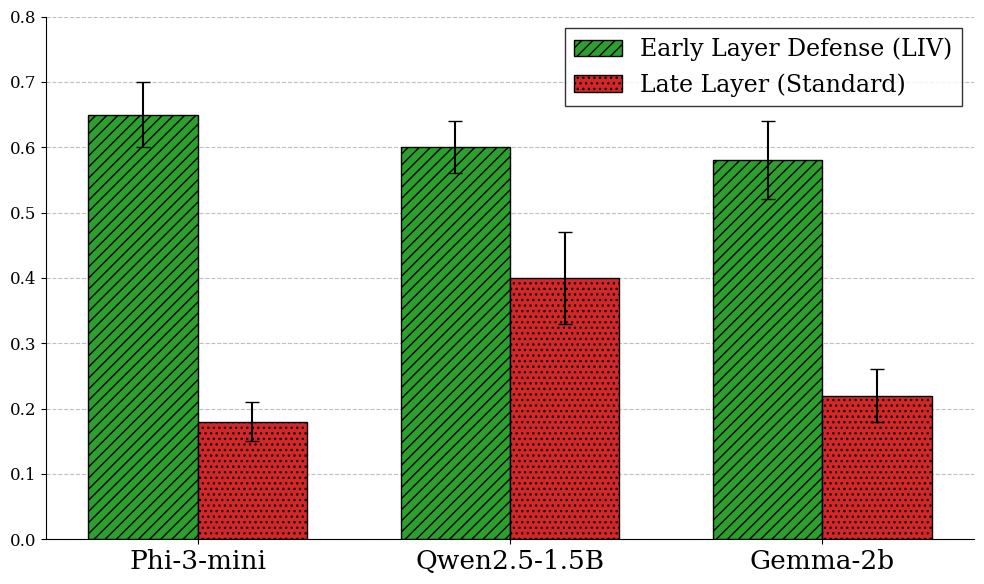}
    \caption{\textbf{The Safety Gap.} Detection rates of adversarial jailbreaks across three model families. Green bars (LIV) consistently outperform Red bars (Standard), highlighting the vulnerability of output-based defenses.}
    \label{fig:safety_gap}
\end{figure}

The primary experiment evaluated the robustness of Latent Intent Verification (LIV) against zero-day semantic camouflage. Figure \ref{fig:safety_gap} illustrates the detection rates for three distinct SLM architectures on the held-out adversarial dataset.

The results reveal a substantial Safety Gap ($\Delta_S$). Standard late-layer defenses (Red bars) consistently failed to detect the majority of camouflaged attacks, achieving detection rates as low as 18\% for Phi-3 and 22\% for Gemma-2b. This confirms that perimeter-based guardrails are effectively "blind" to semantic wrapping.

In sharp contrast, the early-layer LIV probes (Green bars) maintained robust detection rates ranging from 58\% to 65\%. For the Qwen2.5-1.5B model, shifting the defense depth from the final layer to Layer 4 resulted in a relative performance improvement of \textbf{50\%}. This indicates that the "harm signature" is not absent from the model; it is merely suppressed by the subsequent processing layers.

\subsection{Mapping the Intent Horizon}
\begin{figure}[h]
    \centering
    \includegraphics[width=0.48\textwidth]{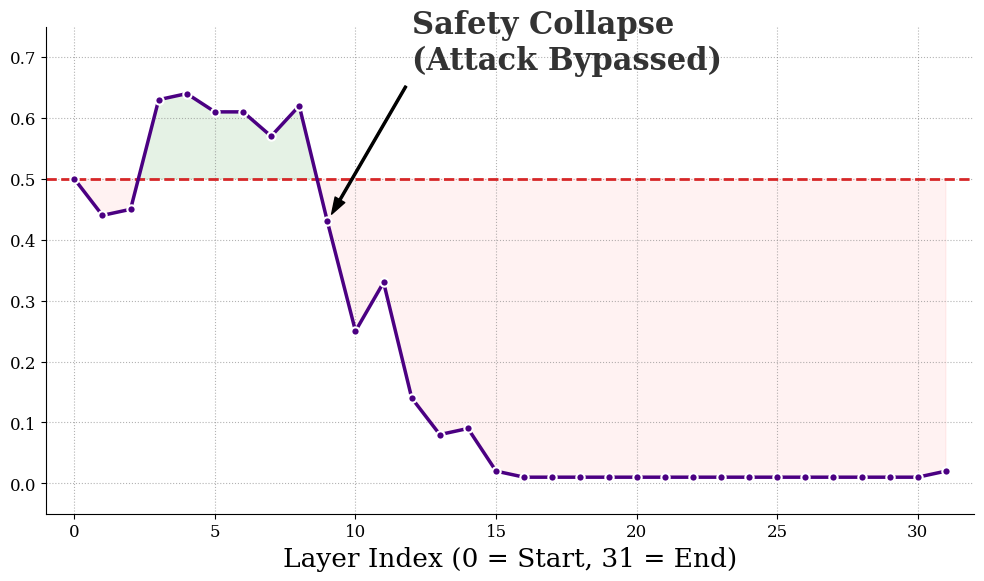}
    \caption{\textbf{The Intent Horizon.} The probability of detecting harmful intent collapses after Layer 10. This visualization proves that safety information is a decaying signal, not a constant one.}
    \label{fig:intent_horizon}
\end{figure}

To understand the dynamics of this suppression, a granular layer-wise sweep was performed on the Microsoft Phi-3 model. Figure \ref{fig:intent_horizon} plots the probability of harm detection $P(Harm)$ against network depth.

The trajectory identifies a clear Collapse Point between Layers 10 and 12.
\begin{itemize}
    \item \textbf{The Truth Zone (Layers 0--10):} The model exhibits high confidence ($>60\%$) in identifying harmful intent, regardless of the benign narrative wrapper.
    \item \textbf{The Camouflage Zone (Layers 12+):} As the model integrates the "movie script" context, the harm probability precipitously drops, flatlining near 0\% at the final output.
\end{itemize}
This trajectory empirically defines the "Intent Horizon"—the depth limit beyond which the model's safety alignment is overridden by its instruction-following capabilities.

\subsection{Geometric Analysis of Latent Space}
To visualize the mechanism of LIV, the hidden states of safe, harmful, and camouflaged queries were projected into 2D space. Table \ref{tab:latent_analysis} summarizes the geometric relationships observed.

At the output layer (Layer $N-1$), the camouflaged attacks are topologically embedded within the "Safe" cluster, making them mathematically indistinguishable from benign queries to any linear classifier. However, at the Intent Horizon (Layer 4), these same attacks appear as distinct outliers, separated from the safe cluster. This \textbf{Geometric Isolation} explains why the simple linear probes of LIV are effective: they operate in a space where the "crime" has not yet been masked by the "alibi."

\begin{table}[h]
\centering
\small
\begin{tabular}{p{2cm} p{2.5cm} p{2cm}}
\toprule
\textbf{Defense Depth} & \textbf{Attack Geometry} & \textbf{Outcome} \\
\midrule
\textbf{Late Layer} (Standard) & Embedded  ``Safe'' Cluster & \textbf{Failure} (False Negative) \\
\midrule
\textbf{Early Layer} (LIV) & Outlier / Near ``Harmful'' Cluster & \textbf{Success} (True Positive) \\
\bottomrule
\end{tabular}
\caption{Geometric analysis of Semantic Camouflage attacks. At early layers, adversarial inputs remain distinct outliers, enabling detection.}
\label{tab:latent_analysis}
\end{table}

\section{Discussion}
\label{sec:discussion}

The evidence presented in this study suggests that it may be time to re-examine how current market-available LLMs are aligned for safety. In this section, the implications of the "Intent Horizon" are interpreted, limitations of the current approach are acknowledged, and directions for future research are proposed.

\subsection{The Illusion of Safety: Permission Structures}
The findings suggest that current Reinforcement Learning from Human Feedback (RLHF) techniques primarily optimize the \textit{surface realization} of refusal. They teach the model \textit{what not to say}, but they do not remove the model's internal understanding of the harmful concept. When a user employs semantic camouflage, they effectively construct a "permission structure"---a narrative context that overrides the surface-level refusal training.

LIV succeeds because it interrogates the model's fundamental understanding before this permission structure is fully processed. This implies that true safety cannot be achieved by merely scaling RLHF on diverse datasets; it requires architectural interventions that enforce consistency between the model's latent "intent" (early layers) and its final output.

\subsection{Limitations}
While LIV demonstrates robust detection capabilities, several limitations are acknowledged:
\begin{itemize}
    \item \textbf{Inference Latency:} Although LIV is lightweight compared to a full "Shadow LLM," probing intermediate layers introduces a non-trivial computational overhead during inference, potentially impacting real-time applications.
    \item \textbf{Adaptive Attacks:} As with any static defense, it is theoretically possible for an adversary to optimize against the specific layer used by LIV (e.g., by using gradient-based attacks to suppress the harm signal at Layer 4). A dynamic probing depth may be required to counter such adaptive adversaries.
    \item \textbf{Model Specificity:} While LIV was validated across three architectures, the precise depth of the "Intent Horizon" varies by model size and training data. Deployment requires a calibration phase to identify the optimal probing depth for each specific model.
\end{itemize}

\subsection{Future Work}
Future research aims to focus on Dynamic Depth Probing, where the defense mechanism adaptively selects which layer to interrogate based on the perplexity or complexity of the input prompt. Additionally, investigations will explore 'Steering Vectors'---not just detecting the harmful intent at Layer 4, but actively intervening to suppress the activation vector, thereby "healing" the model's thought process in real-time.

\section{Conclusion}
\label{sec:conclusion}

This study exposes a critical disconnect in the architecture of Small Language Models: the divergence between \textit{latent cognition} and \textit{surface generation}. It has been demonstrated that safety fine-tuning is insufficient to erase the fundamental knowledge structures acquired during pretraining; while it effectively suppresses explicit threats at the output layer, it leaves the model's pre-trained, early-layer understanding of harmful concepts intact. This persistence of latent knowledge creates an ``Intent Horizon''—a structural vulnerability where semantic camouflage can manipulate the model's context processing to bypass refusal mechanisms.

The proposed defense, \textbf{Latent Intent Verification (LIV)}, validates that the most reliable signal of safety lies not in what the model \textit{says}, but in what it \textit{thinks}. By shifting the defensive perimeter from the output to the latent interior, a 20--50\% improvement in detecting zero-day attacks was achieved.

It is concluded that as LLMs evolve into autonomous agents, external guardrails will become insufficient. True alignment requires {White-Box Safety—architectures where the model's internal state is transparently verifiable before any action is taken. This research provides the first empirical roadmap for measuring and enforcing that transparency.

\bibliographystyle{IEEEtran}
\bibliography{Reference}
\end{document}